\documentclass[twocolumn]{bytedance_seed}
\usepackage{graphicx} 
\usepackage{amsmath}
\usepackage{amssymb}
\usepackage{algorithm}
\usepackage{algorithmic}

\usepackage{newfloat}
\usepackage{listings}
\DeclareCaptionStyle{ruled}{labelfont=normalfont,labelsep=colon,strut=off} 
\floatstyle{ruled}
\newfloat{listing}{tb}{lst}{}
\floatname{listing}{Listing}

\title{ReTrace: Rejected-Trajectory Conditioning for Speculative Decoding}
\author[1,2,*]{Luxi~Lin}
\author[1,*]{Zhanpeng~Zeng}
\author[2,*]{Shuang~Peng}
\author[2,\dagger\ddagger]{Songwei~Liu}
\author[1,\dagger]{Rongrong~Ji}
\affiliation[1]{Xiamen University}
\affiliation[2]{ByteDance}
\contribution[*]{Equal contribution}
\contribution[\dagger]{Corresponding authors}
\contribution[\ddagger]{Tech Lead}
\date{\today}
\correspondence{
    Songwei Liu at \email{21831068@zju.edu.cn},
    Rongrong Ji at \email{rrji@xmu.edu.cn}
}

\begin{document}

\abstract{
Speculative decoding accelerates autoregressive language model inference by having a lightweight draft model propose multiple candidate tokens, which are then verified in parallel by a larger target model.
However, after the first rejection, standard prefix-based verification discards the remaining draft suffix, so the computation spent generating and verifying those positions does not contribute to decoding progress.
Focusing on DFlash, we show that rejected positions in a rejected suffix may still align with the target continuation, indicating that the draft model can retain useful semantic and structural information despite local token-level errors. 
Motivated by this observation and inspired by conditional diffusion, we introduce~\textbf{ReTrace}, a rejected-trajectory conditioning method that conditions each draft block on the rejected suffix from the previous round rather than generating it from fresh mask placeholders alone.
ReTrace retains the hidden representations of the rejected suffixes, aligns them with the next draft block, refines them using target-aware correction signals from the same verification pass, and admits them into the drafter's input embeddings through gated residual fusion.
Because rejected tokens are never committed and target-side verification remains unchanged, ReTrace preserves the lossless property of speculative decoding without requiring an additional model forward pass.
Experiments with Qwen3 models across mathematical reasoning, code generation, and open-ended dialogue demonstrate that ReTrace consistently improves average acceptance length and end-to-end decoding speed over its DFlash backbone. By introducing cross-round conditioning without modifying within-round proposal generation, ReTrace is largely orthogonal to existing drafting improvements and might be combined with them for further gains.
}

\maketitle


\section{Introduction}

The inference efficiency of large language models~\cite{radford2018gpt,touvron2023llama,bai2023qwen} is fundamentally constrained by autoregressive decoding. During standard generation, each token requires a new forward pass conditioned on the entire previously generated context, causing latency to grow with output length. This sequential process is particularly costly for long-form tasks such as mathematical reasoning, code generation, multi-turn dialogue, and agentic workflows, and it also prevents modern GPUs from fully exploiting their parallel computing capacity. Reducing sequential computation during decoding without degrading generation quality has therefore become a central problem in efficient large language model inference.

\begin{figure*}[t]
\centering
\includegraphics[width=0.85\textwidth]{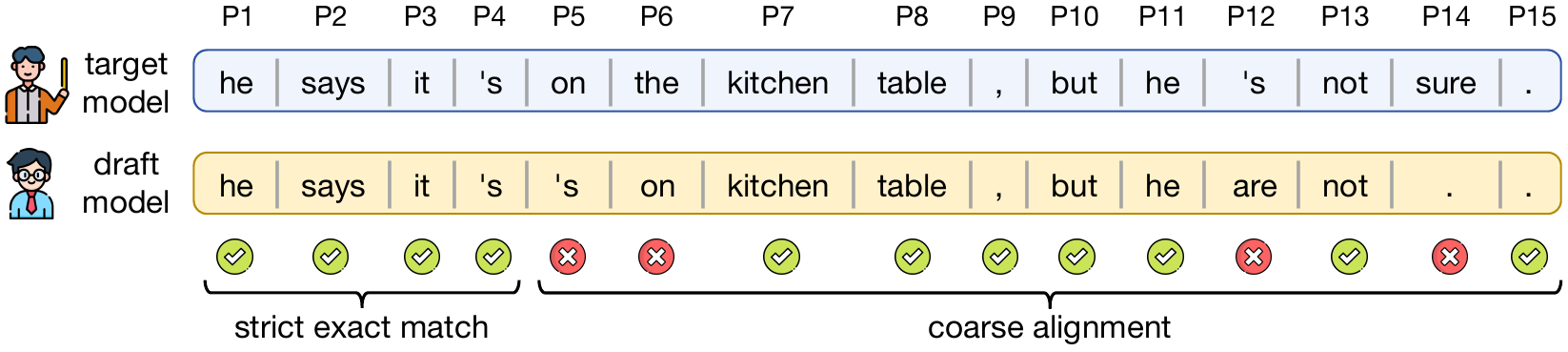}
\caption{An MT-Bench example with Qwen3-4B as the target model. Although verification stops at the first mismatch, later draft tokens realign with the target, suggesting that the rejected suffix retains useful predictive information.}
\label{fig:example}
\end{figure*}

Speculative decoding~\cite{leviathan2023fast} addresses this bottleneck through a draft-then-verify paradigm. A lightweight draft model first proposes multiple candidate tokens, which are then verified in parallel by a larger target model. When the proposed sequence agrees with the target model, multiple tokens can be committed in a single verification step, reducing the number of expensive target-model forward passes. Under standard verification, the final output remains controlled by the target model. However, the resulting speedup depends heavily on draft–target agreement. A long accepted prefix allows each verification step to advance multiple tokens, whereas an early mismatch causes the remaining candidates to be rejected and prevents the computation spent on drafting them from translating into actual decoding progress.


Existing studies primarily improve draft quality while controlling drafting overhead through better drafter architectures, candidate organization, and verification strategies~\cite{cai2024medusa,li2024eagle,chen2026dflash,liu2025pearl}. Representative directions include multi-head prediction~\cite{cai2024medusa}, feature-level drafting~\cite{li2024eagle}, and tree-structured candidate construction~\cite{li2024eagle2}. More recently, diffusion-based parallel drafters such as DFlash~\cite{chen2026dflash} generate an entire candidate block in a single drafting pass without autoregressively conditioning on previously sampled draft tokens, achieving strong end-to-end speedups.

Despite these advances, a fundamental inefficiency remains in the draft-then-verify paradigm: each draft is used only for its current verification round. After the first mismatch, the remaining tokens and their hidden representations are discarded, even though the computation required to produce and verify them has already been incurred. This raises a natural question: does the rejected computation retain information about the target continuation, and can it be reused to condition the next draft?

In this work, we study this question using DFlash, whose diffusion-based parallel decoding provides a clean setting for examining information retained beyond the first mismatch. Figure~\ref{fig:example} provides an illustrative example. Verification terminates at the first mismatch, resulting in a short
accepted prefix. Nevertheless, despite local repetitions, omissions, and incorrect predictions, later draft positions realign with the target continuation. 
This qualitative observation suggests that an early token-level error does not necessarily render all subsequent draft computation uninformative. Beyond this example, our quantitative analysis further shows that rejected-suffix representations retain predictive information about the target continuation.

To this end, we introduce \textbf{ReTrace}, a rejected-trajectory conditioning method for speculative decoding. ReTrace preserves the standard draft-then-verify paradigm while extending information flow across rounds. Instead of discarding the rejected trajectory after verification, it carries the corresponding hidden representations forward as auxiliary conditioning for the next proposal. Since all required draft and target representations are already available from standard drafting and verification, ReTrace requires no additional forward pass through either model.
Our contributions are twofold. First, we identify and quantitatively analyze the predictive signal retained in rejected trajectories of DFlash, showing that representations beyond the first mismatch can remain informative about the target continuation. Second, we introduce ReTrace, a rejected-trajectory conditioning method that extends information flow across drafting rounds and improves speculative decoding with marginal drafting overhead.

\section{Related Works}

\paragraph{Speculative Decoding.}
Speculative decoding~\cite{leviathan2023fast} uses a lightweight drafter to propose multiple tokens that are verified in parallel by a target model, reducing expensive target-model forward passes.
Recent work improves draft quality and efficiency through feature-level prediction (EAGLE~\cite{li2024eagle,li2025eagle3}), single-pass block diffusion (DFlash~\cite{chen2026dflash}), intra-block causal refinement (Domino~\cite{huang2026domino}), confidence-scheduled semi-autoregressive drafting (DSpark~\cite{cheng2026dspark}), and parallel tree drafting (JetFlow~\cite{hu2026jetspec}).
Unlike these methods, which strengthen proposal generation or verification within the current round, ReTrace carries rejected-suffix representations across rounds, refines them with target states from verification, and conditions the next proposal on them. It is therefore complementary to existing improvements.

\paragraph{Draft Reuse.}
Standard prefix verification discards the remaining draft suffix after the first rejection, motivating methods that reuse otherwise wasted draft computation. Token Recycling~\cite{luo2025turning} caches discrete candidate-token transitions to construct future draft trees. More closely related to our work, Make Every Draft Count~\cite{chen2026make} introduces Lyanna, which fundamentally redesigns the drafter to autoregressively generate token-independent hidden-state trajectories, postpones token conditioning until sampling, and resamples candidate trees from the same hidden states after verification failures. The resampled candidates must then be verified together with a subsequent regular draft.


ReTrace differs from Lyanna in how the discarded computation is reused. Lyanna resamples candidate trees from reusable hidden trajectories after a verification failure, and these candidates must still undergo target-model verification. ReTrace instead uses the rejected states as conditioning signals for the next regular draft. Thus, the reused states need not themselves generate valid candidates; they only need to provide predictive information that improves subsequent drafting.

\section{Predictive Signal in Rejected Trajectories}
\label{sec:suffix_analysis}

We ask whether the suffix strictly after the first rejected position retains predictive information about the target continuation. 
Using Qwen3-4B with its DFlash-b16 drafter under greedy decoding, we quantify the retained predictive signal in rejected suffixes from two perspectives: whether the target token remains near the top of the draft distribution, and whether the rejected draft states remain aligned with the corresponding target hidden trajectory.

\paragraph{Choice of Drafter.}
We focus on DFlash~\cite{chen2026dflash} because it provides a cleaner source of rejected trajectories for cross-round conditioning. First, in EAGLE-style autoregressive drafters~\cite{li2024eagle,li2025eagle3}, later draft states are generated from earlier sampled tokens or features. An early error therefore changes the conditioning context of all descendant states and can propagate through the remainder of the branch. DFlash instead predicts a block in parallel: later positions may interact, but they are not generated autoregressively from the sampled token at the preceding position. A local rejection therefore does not necessarily invalidate all subsequent states.

Second, EAGLE distributes much of its drafting and verification computation across a shallow tree of alternative branches. Once one branch is selected, states on the remaining branches represent alternative continuations from earlier decision points rather than consecutive future positions along the realized continuation, making them difficult to reuse as position-aligned conditioning. DFlash instead produces a single, deeper block, so a rejection can leave a longer sequence of hidden states naturally aligned with subsequent draft positions. These properties make DFlash a particularly suitable setting for studying rejected-trajectory conditioning.

\paragraph{Token-Level Predictive Signal.}
Our first probe asks whether the target token remains plausible under the draft distribution at positions beyond the first mismatch. Although these positions no longer contribute to prefix acceptance, their draft distributions may still contain information about the target continuation. We measure this information by recording the rank of the target token at each suffix position.
Specifically, for a suffix position \(j\), let \(z^D_{j,v}\) be the draft logit for vocabulary token \(v\) and let \(y^T_j\) be the token that the target model actually produces at that position. Its rank over the vocabulary \(\mathcal{V}\) is
\begin{equation}
\operatorname{rank}(y^T_j)
=
1+
\sum_{v\in\mathcal{V}}
\mathbf{1}\!\left[z^D_{j,v}>z^D_{j,y^T_j}\right].
\label{eq:target_token_rank}
\end{equation}
Rank one means that the target token is the drafter’s top prediction, while a small rank means that only a few vocabulary tokens receive higher draft scores.

\begin{figure}[t]
\centering
\includegraphics[width=\columnwidth]{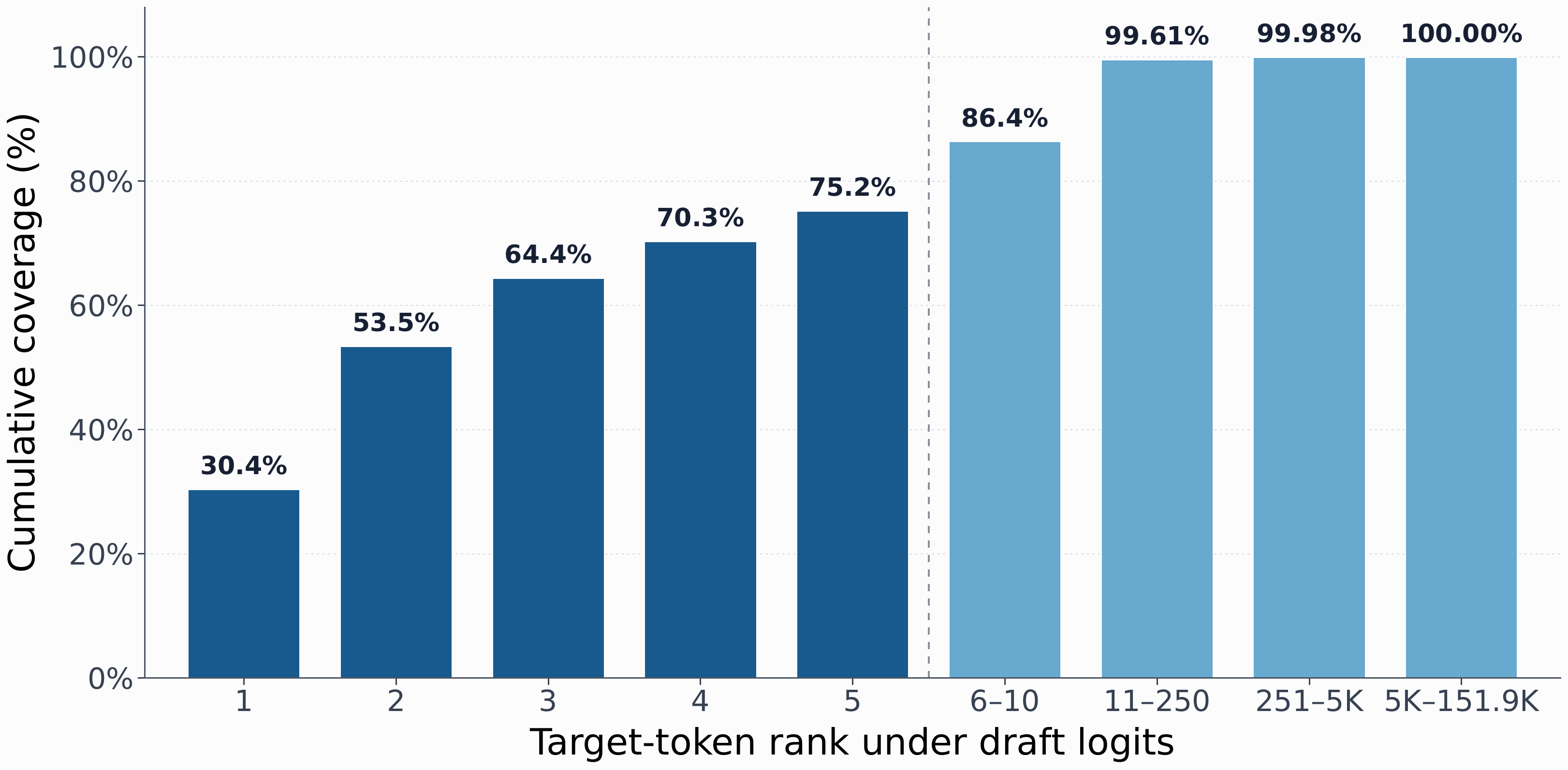}
\caption{Cumulative coverage of the true target token's rank under the base
DFlash logits. The first rejected proposal is excluded, and ranks are
computed over the complete 151,936-token vocabulary. Each bar gives the
fraction of analyzed rejected-suffix positions whose target token falls at or
below the indicated rank range.}
\label{fig:target_token_rank}
\end{figure}

Figure~\ref{fig:target_token_rank} shows that \(30.4\%\) of target tokens remain the drafter's top prediction, while \(75.2\%\) fall within its top five. Thus, even after an earlier mismatch, the target token often remains among a small set of high-scoring alternatives, indicating that the rejected suffix retains coarse token-level predictive information.

\paragraph{Representation-Level Predictive Signal.}
Token ranks characterize the drafter’s output distribution. We next ask whether the rejected hidden states retain information specific to the current continuation. For this offline analysis, we replay the target model’s actual continuation and collect its hidden states at the same relative positions. We then compute the similarity between each rejected draft state and the corresponding target hidden state. For a block \(b\) with a suffix of length \(L_b\), the same-position trajectory similarity is
\begin{equation}
S_b^{+}
=
\frac{1}{L_b}
\sum_{j=1}^{L_b}
\cos\!\left(h^D_{b,j},h^{T,+}_{b,j}\right),
\label{eq:true_trajectory_similarity}
\end{equation}
where \(h^D_{b,j}\) is a rejected draft state and \(h^{T,+}_{b,j}\) its counterpart on the same sample's true target continuation. To distinguish continuation-specific alignment from generic hidden-space similarity, we additionally introduce a random baseline: we keep the rejected draft states fixed and replace the true target continuation with target hidden-state windows drawn from other sequences.

\begin{table}[t]
\centering
\footnotesize
\setlength{\tabcolsep}{1pt}
\renewcommand{\arraystretch}{1.05}
\resizebox{\columnwidth}{!}{%
\begin{tabular}{lccccc}
\toprule
Comparison & AIME25 & GSM8K & HumanEval & MT-Bench & Overall \\
\midrule
True trajectory & 0.396 & 0.330 & 0.307 & 0.371 & 0.351 \\
Random  & 0.171 & 0.121 & 0.105 & 0.115 & 0.128 \\
Gap             & +0.225 & +0.209 & +0.202 & +0.256 & +0.223 \\
\bottomrule
\end{tabular}%
}
\caption{Cosine similarity between rejected draft hidden states and target
hidden trajectories. Random target windows are drawn from other sequences in
the same benchmark and matched by suffix length and generation progress.
Overall is the macro-average across benchmarks.}
\label{tab:hidden_trajectory_similarity}
\end{table}

As shown in Table~\ref{tab:hidden_trajectory_similarity}, rejected draft states are substantially more similar to their own target continuation than to random windows from other samples. 
The gap is positive on all four benchmarks, from \(0.202\) on HumanEval to \(0.256\) on MT-Bench, and the macro-average similarity rises from \(0.128\) to \(0.351\). 
These results provide strong evidence that rejected trajectories retain continuation-specific information. At the same time, the moderate absolute similarity indicates that their alignment with the target trajectory remains imperfect.

Together, the two probes establish a bounded form of reusability. The token-level analysis shows that the target token often remains among a small set of high-scoring alternatives, while the representation-level analysis shows that continuation-specific information persists in the rejected hidden states. The rejected suffix is therefore not reliable enough to accept or copy directly, but can provide auxiliary conditioning for subsequent drafting.

\section{Method}

\begin{figure*}[t]
\centering
\includegraphics[width=0.89\textwidth]{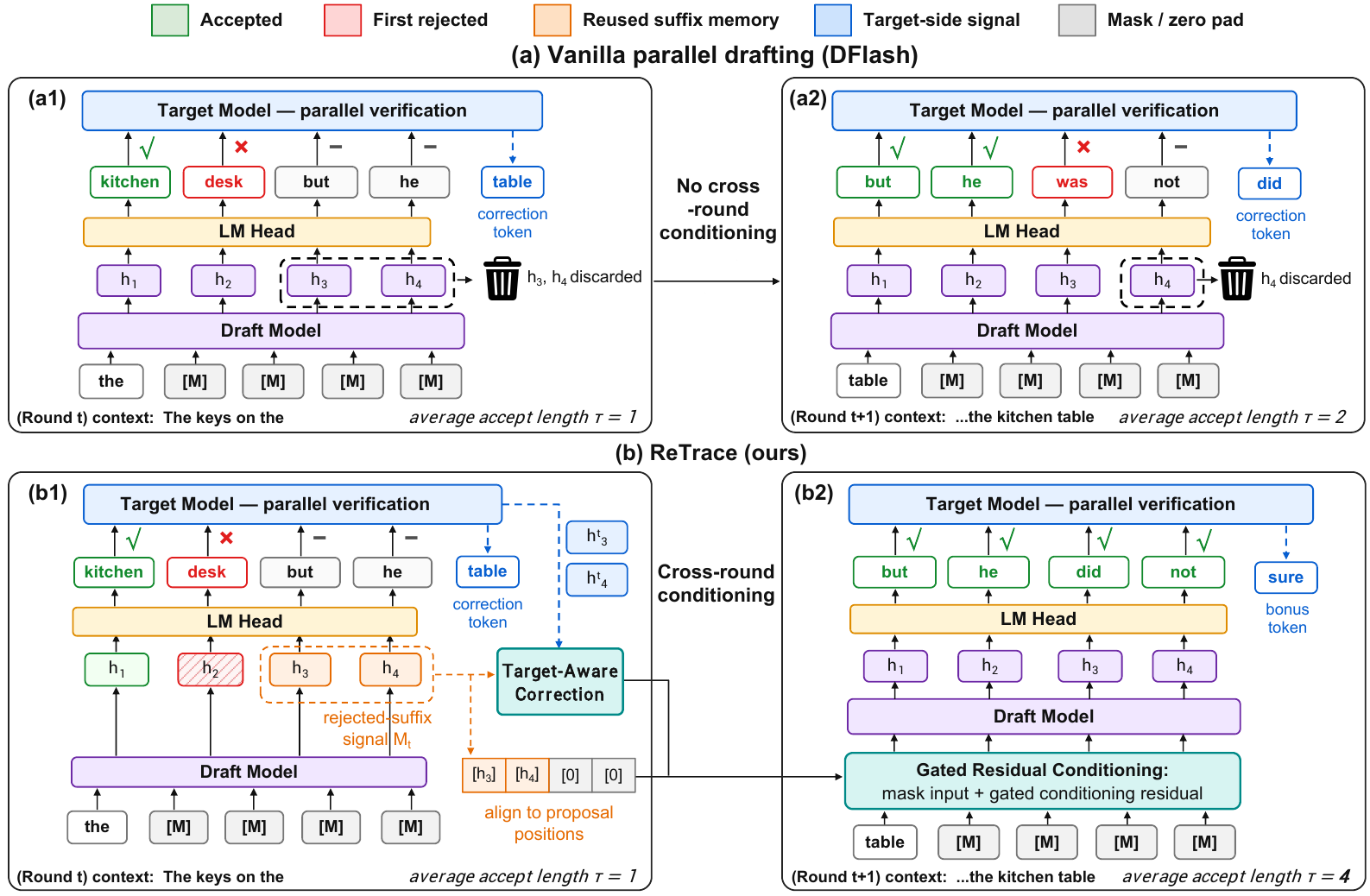}
\caption{Illustration of ReTrace. The target continuation is ``The keys on the
kitchen table, but he did not find them.'' After drafting ``kitchen desk but
he'' from the context ``The keys on the'', the verifier accepts ``kitchen'',
rejects ``desk'', and provides ``table'' as the target-side correction.
Vanilla speculative drafting discards the hidden states of the remaining
rejected suffix (``but he''), whereas ReTrace aligns them with the next block and
conditions it on them after target-guided correction and gated residual fusion.}
\label{fig:framework}
\end{figure*}

\paragraph{Overview.}
The preceding analysis shows that the rejected suffix contains useful information, although its tokens are not reliable enough to accept directly. Inspired by conditional diffusion, ReTrace treats the corresponding hidden-state sequence as a rejected trajectory and uses it as auxiliary conditioning for the next masked draft block. This extends DFlash from conditioning each block only on the verified context to conditioning it also on information carried across drafting rounds.

Concretely, ReTrace injects the hidden representations of the rejected suffix into the drafter's input embeddings through a learned gate. 
More specifically, ReTrace retains and aligns the rejected-suffix representations with the next proposal positions. It then fuses each aligned draft state with the corresponding target representation already available from verification, providing additional target-side information. The resulting target-conditioned trajectory is injected into the next-round mask inputs through a learned gate.
The accepted prefix and the first rejected position are excluded because both have already been resolved by verification, with the latter replaced by the target-side correction token. 
The conditioning lasts for one round, while the target model, verification rule, and token acceptance
procedure remain unchanged.
We elaborate on each component in the following text.

\paragraph{Rejected-Trajectory Construction and Alignment.}
At decoding round $t$, the drafter produces a block of $B$ candidate tokens together with their final-layer representations,
\begin{equation}
\left(
h^{D}_{t,1},
h^{D}_{t,2},
\ldots,
h^{D}_{t,B}
\right).
\end{equation}
Let $A_t$ denote the number of draft tokens accepted by the target model. When $A_t<B$, the first rejected position is $r_t=A_t+1$. 
ReTrace excludes the accepted-prefix representations $h^D_{t,1:A_t}$, since their corresponding tokens have already been committed to the verified context. It also excludes $h^D_{t,r_t}$, because
the position is resolved by the target-side correction token. This information is available to the next drafting round through the standard context pathway.


ReTrace therefore carries forward only the representations strictly after the first rejected position:
\begin{equation}
\left(
h^{D}_{t,r_t+1},
h^{D}_{t,r_t+2},
\ldots,
h^{D}_{t,B}
\right).
\label{eq:rejected_memory}
\end{equation}
These representations are not assumed to form a correct continuation. They serve as the conditioning signal for the upcoming drafting block, one that may preserve the semantic direction, formatting pattern, or local structure predicted by the drafter.

To use this signal in round $t+1$, ReTrace aligns it with the proposal positions of the next draft block. After the target model commits the correction token at the rejection boundary, $h^{D}_{t,r_t+1}$ corresponds to the first unresolved position in the next round. We therefore define the aligned signal as
\begin{equation}
m_{t+1,j}
=
\begin{cases}
h^{D}_{t,r_t+j}, & 1 \leq j \leq B - r_t, \\
\mathbf{0}, & B - r_t < j \leq B,
\end{cases}
\label{eq:trajectory_alignment}
\end{equation}
This alignment shifts the retained states to the beginning of the next proposal block while preserving their original order. 
If the full block is accepted, or if no representation remains after excluding the first rejected position, then ReTrace reduces to the original drafting procedure.

\paragraph{Target-Aware Correction.}
The aligned rejected trajectory retains predictive information, but it reflects only the drafter’s prediction and may carry errors from the failed draft path. The target representations produced during verification provide a complementary view of how the target model evaluates the same trajectory. ReTrace therefore adjusts each aligned draft state with its corresponding target representation to form a target-corrected trajectory, without requiring an additional target-model forward pass.


For each aligned position $j$, let $u_{t+1,j}$ denote the target representation whose output logits score the candidate token at position $r_t+j$. In a causal model the state at position $r_t+j-1$ is the one that predicts the token at position $r_t+j$, so
\begin{equation}
u_{t+1,j}
=
\begin{cases}
h^{T}_{t,r_t+j-1}, & 1 \leq j \leq B - r_t, \\
\mathbf{0}, & B - r_t < j \leq B,
\end{cases}
\label{eq:target_state_alignment}
\end{equation}
where $h^{T}_{t,i}$ is the final-layer target representation produced during verification.
Since the two representations come from different models, we let a lightweight projection map the pair into an update in the draft representation space:
\begin{equation}
c_{t+1,j} = W_c \left[m_{t+1,j}; u_{t+1,j}\right],
\label{eq:memory_correction_update}
\end{equation}
where $W_c\in\mathbb{R}^{d\times 2d}$ is learnable. The corrected signal is then
\begin{equation}
r_{t+1,j}
=
m_{t+1,j}
+
\beta\,c_{t+1,j}.
\label{eq:corrected_memory}
\end{equation}
The residual formulation preserves the original suffix representation while allowing the correction module to make a target-aware adjustment. Here, $\beta$ controls the adjustment strength and is gradually increased during the initial training stage for optimization stability. At inference time, the fully activated coefficient is used.

\subsection{Gated Residual Conditioning}
\label{sec:gated_memory_injection}

The corrected signal carries information from the previous drafting round, but it should not replace the current mask input. ReTrace therefore admits it as a gated residual. Let $e_{t+1,j}$ denote the drafter's input representation at proposal position $j$ of round $t+1$, with $j=0$ the verified anchor. For each conditioned position, the conditioned input is
\begin{equation}
\widetilde{e}_{t+1,j}
=
e_{t+1,j}
+
W_v r_{t+1,j}
\odot
\sigma\!\left(
W_g\left[e_{t+1,j};r_{t+1,j}\right]
\right),
\label{eq:gated_conditioning}
\end{equation}
where $W_v\in\mathbb{R}^{d\times d}$ and $W_g\in\mathbb{R}^{d\times 2d}$ are learnable, $\sigma$ is the element-wise sigmoid, and $\odot$ denotes element-wise multiplication. The gate is vector-valued, so it controls how much of the signal enters each hidden dimension. The anchor and the positions without an aligned signal keep their original inputs, $\widetilde{e}_{t+1,j}=e_{t+1,j}$.

The original draft input therefore remains the main pathway, and the conditioning signal enters only as a residual that the gate admits selectively. The conditioned inputs are passed through the original drafter to produce the next candidate block, and ReTrace introduces no further changes to the remaining draft computation. For stable optimization, $W_v$ is initialized to zero, so the residual vanishes at initialization and the drafter starts from its original behavior, gradually learning to condition on the rejected suffix when it benefits subsequent proposals.

\subsection{Optimization and Inference}
\label{sec:optimization_and_inference}

During training, cached draft and target representations are detached before being reused in the next decoding round, preventing gradients from propagating across rounds. The draft backbone, correction module, and gated injection module are jointly optimized using the current-round prediction loss, while the target model remains frozen.

At inference time, ReTrace reuses representations already produced by the standard drafting and verification passes. Each signal is retained for a single round and then replaced by the one derived from the latest rejected suffix. ReTrace introduces no additional target-model forward pass and leaves the target-side acceptance and output commitment procedures unchanged.

\section{Experiments}

\begin{table*}[!t]
    
    \centering
    \resizebox{\linewidth}{!}{
    \scriptsize
    \setlength{\tabcolsep}{1.2pt}
    \begin{tabular}{
        c l
        @{\hspace{1.0em}} cc cc cc
        @{\hspace{1.0em}} cc cc
        @{\hspace{1.0em}} cc cc
        @{\hspace{1.0em}} cc
    }
        \toprule
        \multirow{2}{*}{Model}
        & \multirow{2}{*}{Method}
        & \multicolumn{6}{c}{\textsc{Math}}
        & \multicolumn{4}{c}{\textsc{Code}}
        & \multicolumn{4}{c}{\textsc{Chat}}
        & \multicolumn{2}{c}{\textsc{Overall}} \\
        \cmidrule(lr){3-8}
        \cmidrule(lr){9-12}
        \cmidrule(lr){13-16}
        \cmidrule(lr){17-18}

        & 
        & \multicolumn{2}{c}{GSM8K}
        & \multicolumn{2}{c}{MATH-500}
        & \multicolumn{2}{c}{AIME25}
        & \multicolumn{2}{c}{HumanEval}
        & \multicolumn{2}{c}{LCB}
        & \multicolumn{2}{c}{MT-Bench}
        & \multicolumn{2}{c}{Alpaca}
        & \multicolumn{2}{c}{\textit{Avg.}} \\
        \midrule

        \multicolumn{2}{c}{Temperature $=0$}
        & Speedup & $\tau$
        & Speedup & $\tau$
        & Speedup & $\tau$
        & Speedup & $\tau$
        & Speedup & $\tau$
        & Speedup & $\tau$
        & Speedup & $\tau$
        & Speedup & $\tau$ \\
        \midrule

        \multirow{4}{*}{Qwen3-4B}
        & EAGLE-3 (16)
        & $1.67\times$ & 3.31 & $1.67\times$ & 3.14 & $1.76\times$ & 3.25
        & $1.58\times$ & 3.07 & $1.74\times$ & 3.03
        & $1.42\times$ & 2.91 & $1.43\times$ & 2.88 & $1.61\times$ & 3.08 \\

        & EAGLE-3 (60)
        & $1.67\times$ & 3.70 & $1.67\times$ & 3.49 & $1.71\times$ & 3.61
        & $1.61\times$ & 3.37 & $1.81\times$ & 3.35
        & $1.43\times$ & 3.21 & $1.42\times$ & 3.17 & $1.62\times$ & 3.41 \\

        & DFlash (16)
        & $3.37\times$ & 6.36 & $5.18\times$ & 8.25 & $5.32\times$ & 8.66
        & $4.15\times$ & 6.69 & $4.68\times$ & 7.35
        & $1.85\times$ & 4.07 & $1.66\times$ & 3.48 & $3.74\times$ & 6.41 \\

        & ReTrace (16)
        & $\mathbf{3.73\times}$ & \textbf{7.36}
        & $\mathbf{5.47\times}$ & \textbf{8.88}
        & $\mathbf{5.44\times}$ & \textbf{8.94}
        & $\mathbf{4.27\times}$ & \textbf{6.90}
        & $\mathbf{4.85\times}$ & \textbf{7.80}
        & $\mathbf{1.99\times}$ & \textbf{4.40}
        & $\mathbf{1.73\times}$ & \textbf{3.69}
        & $\mathbf{3.93\times}$ & \textbf{6.85} \\

        \midrule

        \multirow{4}{*}{Qwen3-8B}
        & EAGLE-3 (16)
        & $1.85\times$ & 3.24 & $2.04\times$ & 3.15 & $2.23\times$ & 3.65
        & $1.93\times$ & 3.21 & $1.86\times$ & 2.86
        & $1.67\times$ & 2.73 & $1.50\times$ & 2.60 & $1.87\times$ & 3.06 \\

        & EAGLE-3 (60)
        & $1.92\times$ & 3.61 & $2.18\times$ & 3.52 & $2.06\times$ & 3.65
        & $1.96\times$ & 3.55 & $1.86\times$ & 3.14
        & $1.71\times$ & 3.01 & $1.53\times$ & 2.80 & $1.89\times$ & 3.33 \\

        & DFlash (16)
        & $3.91\times$ & 6.40 & $5.89\times$ & 8.21 & $5.96\times$ & 8.40
        & $4.03\times$ & 6.49 & $4.26\times$ & 7.17
        & $2.39\times$ & 4.18 & $1.89\times$ & 3.59 & $4.05\times$ & 6.35 \\

        & ReTrace (16)
        & $\mathbf{4.34\times}$ & \textbf{7.39}
        & $\mathbf{6.10\times}$ & \textbf{8.95}
        & $\mathbf{6.10\times}$ & \textbf{8.73}
        & $\mathbf{4.21\times}$ & \textbf{6.74}
        & $\mathbf{4.38\times}$ & \textbf{7.50}
        & $\mathbf{2.47\times}$ & \textbf{4.41}
        & $\mathbf{1.97\times}$ & \textbf{3.81}
        & $\mathbf{4.22\times}$ & \textbf{6.79} \\

        \midrule

        \multicolumn{2}{c}{Temperature $=1$}
        & Speedup & $\tau$
        & Speedup & $\tau$
        & Speedup & $\tau$
        & Speedup & $\tau$
        & Speedup & $\tau$
        & Speedup & $\tau$
        & Speedup & $\tau$
        & Speedup & $\tau$ \\
        \midrule

        \multirow{4}{*}{Qwen3-4B}
        & EAGLE-3 (16)
        & $1.51\times$ & 3.26 & $1.38\times$ & 3.03 & $1.35\times$ & 2.86
        & $1.39\times$ & 3.02 & $1.34\times$ & 2.99
        & $1.30\times$ & 2.89 & $1.28\times$ & 2.82 & $1.36\times$ & 2.98 \\

        & EAGLE-3 (60)
        & $1.49\times$ & 3.63 & $1.36\times$ & 3.38 & $1.30\times$ & 3.14
        & $1.37\times$ & 3.30 & $1.37\times$ & 3.28
        & $1.24\times$ & 3.14 & $1.25\times$ & 3.11 & $1.34\times$ & 3.28 \\

        & DFlash (16)
        & $2.84\times$ & 6.00 & $3.23\times$ & 6.99 & $2.46\times$ & 5.18
        & $2.96\times$ & 6.04 & $2.92\times$ & 7.04
        & $1.59\times$ & 3.96 & $\mathbf{1.45\times}$ & 3.43 & $2.49\times$ & 5.52 \\

        & ReTrace (16)
        & $\mathbf{3.15\times}$ & \textbf{6.82}
        & $\mathbf{3.39\times}$ & \textbf{7.50}
        & $\mathbf{2.60\times}$ & \textbf{5.41}
        & $\mathbf{2.99\times}$ & \textbf{6.30}
        & $\mathbf{3.56\times}$ & \textbf{7.60}
        & $\mathbf{1.62\times}$ & \textbf{4.16}
        & $1.44\times$ & \textbf{3.56}
        & $\mathbf{2.68\times}$ & \textbf{5.91} \\

        \midrule

        \multirow{4}{*}{Qwen3-8B}
        & EAGLE-3 (16)
        & $1.69\times$ & 3.14 & $1.60\times$ & 2.94 & $1.52\times$ & 2.76
        & $1.68\times$ & 3.10 & $1.47\times$ & 2.82
        & $1.42\times$ & 2.67 & $1.36\times$ & 2.55 & $1.53\times$ & 2.85 \\

        & EAGLE-3 (60)
        & $1.73\times$ & 3.51 & $1.58\times$ & 3.28 & $1.51\times$ & 3.06
        & $1.69\times$ & 3.41 & $1.29\times$ & 3.12
        & $1.39\times$ & 2.89 & $1.35\times$ & 2.74 & $1.51\times$ & 3.14 \\

        & DFlash (16)
        & $3.15\times$ & 5.82 & $3.53\times$ & 6.81 & $2.99\times$ & 5.34
        & $3.24\times$ & 5.63 & $3.52\times$ & 7.12
        & $1.76\times$ & 3.73 & $1.63\times$ & 3.44 & $2.83\times$ & 5.41 \\

        & ReTrace (16)
        & $\mathbf{3.48\times}$ & \textbf{6.63}
        & $\mathbf{3.95\times}$ & \textbf{7.36}
        & $\mathbf{3.04\times}$ & \textbf{5.58}
        & $\mathbf{3.32\times}$ & \textbf{5.79}
        & $\mathbf{3.59\times}$ & \textbf{7.40}
        & $\mathbf{1.82\times}$ & \textbf{4.08}
        & $\mathbf{1.68\times}$ & \textbf{3.69}
        & $\mathbf{2.98\times}$ & \textbf{5.79} \\

        \bottomrule
    \end{tabular}
    }
    \caption{
    End-to-end decoding speedup over vanilla autoregressive decoding
    and average acceptance length ($\tau$) on Qwen3 models.
    Parenthesized values denote the draft tree size for EAGLE-3
    and the draft block size for DFlash and ReTrace.
    The average is computed over all benchmarks.
    Bold values indicate the best result within each model and temperature setting.
    }
    \label{tab:main-results}
\end{table*}


\paragraph{Models and Benchmarks.}
We evaluate ReTrace on two target-model scales,
Qwen3-4B and Qwen3-8B~\cite{qwen2025qwen3}, using the
corresponding DFlash-b16 checkpoints~\cite{chen2026dflash}
as the base drafters. Our evaluation covers seven benchmarks
from three representative generation categories. For
mathematical reasoning, we use GSM8K~\cite{cobbe2021training},
MATH~\cite{hendrycks2021measuring}, and
AIME25~\cite{maa2025aime}. For code generation, we evaluate on
HumanEval~\cite{chen2021evaluating} and
LiveCodeBench~\cite{jain2024livecodebench}. For open-ended
dialogue and instruction following, we use
MT-Bench~\cite{zheng2023judging} and
Alpaca~\cite{taori2023alpaca}. Unless otherwise specified,
all methods use a draft block size of 16, and thinking mode is
disabled during evaluation. Training data, optimization
hyperparameters, hardware, and full generation configurations
are deferred to the appendix.

\paragraph{Evaluation Metrics.}
We report average acceptance length $\tau$, the mean number of draft tokens accepted per target verification step, and end-to-end speedup over standard autoregressive decoding.

\paragraph{Baselines.}
We compare ReTrace with EAGLE-3~\cite{li2025eagle3},
a strong feature-level speculative decoding method. Since
ReTrace is implemented on top of DFlash, we also
include the corresponding official DFlash-b16
model~\cite{chen2026dflash} as the direct backbone baseline.
This comparison isolates the improvement introduced by
rejected-suffix conditioning from that of the underlying
parallel drafter. Standard autoregressive decoding is used as
the reference for calculating end-to-end speedup. For each
target-model scale, all applicable methods are evaluated on
the same benchmark prompts with consistent generation
settings.

\subsection{Main Results}

As shown in Table~\ref{tab:main-results}, ReTrace
consistently strengthens its DFlash backbone across both
model scales and decoding regimes. Averaged over the four
model--temperature configurations, it improves the
macro-average acceptance length by $6.97\%$ and the
macro-average speedup by $5.34\%$ over DFlash. The acceptance
gain is consistent across every benchmark and persists under
both greedy and stochastic decoding, suggesting that
rejected-suffix representations provide reusable correction
signals rather than benefiting only a particular model scale
or decoding trajectory. Longer accepted blocks generally
translate into higher end-to-end speedup. The few small
per-benchmark deviations indicate that wall-clock acceleration
also depends on workload length and fixed per-round overhead;
nevertheless, every model--temperature configuration achieves
a higher macro-average speedup.

\subsection{Component Ablation}
\label{sec:component_ablation}

We ablate how the rejected suffix enters the next draft block. All variants build
the same aligned conditioning signal and differ only in how it is used:
overwrite replaces the mask input representations with it, gated
adds it through the gated residual connection while keeping the mask pathway
intact, and corrected applies target-aware delta correction before fusion.

\begin{table*}[t]
\centering
\setlength{\tabcolsep}{6pt}
\renewcommand{\arraystretch}{1.06}
\resizebox{\textwidth}{!}{%
\begin{tabular}{lccccc}
\toprule
Method & GSM8K & HumanEval & AIME25 & MT-Bench & Avg. \\
\midrule
DFlash (no conditioning)
& $6.36 / 3.37\times$
& $6.69 / 4.15\times$
& $8.66 / 5.32\times$
& $4.07 / 1.85\times$
& $6.45 / 3.67\times$ \\
\midrule
$+$ raw conditioning, overwrite
& $6.67 / 3.47\times$
& $6.54 / 4.05\times$
& $8.72 / 5.33\times$
& $4.11 / 1.86\times$
& $6.51 / 3.68\times$ \\
$+$ raw conditioning, gated
& $7.21 / 3.67\times$
& $6.81 / 4.22\times$
& $8.83 / 5.37\times$
& $4.29 / 1.93\times$
& $6.79 / 3.80\times$ \\
$+$ corrected conditioning, gated
& $\mathbf{7.36 / 3.73\times}$
& $\mathbf{6.90 / 4.27\times}$
& $\mathbf{8.94 / 5.44\times}$
& $\mathbf{4.40 / 1.99\times}$
& $\mathbf{6.90 / 3.86\times}$ \\
\bottomrule
\end{tabular}%
}
\caption{Component ablation on Qwen3-4B under greedy decoding. The three
conditioned variants reuse the same aligned conditioning signal and differ only
in how it enters the next draft block. Each entry reports average acceptance
length $\tau$ / end-to-end speedup, and the last column is the arithmetic mean
over the four benchmarks. Bold values indicate the best result in each column.}
\label{tab:component_ablation}
\end{table*}

As shown in Table~\ref{tab:component_ablation}, even direct overwriting yields a
small improvement, raising average acceptance from $6.45$ to $6.51$. This
supports our analysis that rejected-suffix hidden states contain useful
information, while showing that such a simple injection cannot exploit it
effectively. Gated fusion brings clearer gains by preserving the current-round
mask input, reaching $6.79$, and target-aware correction further improves all
four benchmarks, achieving the best overall result of $6.90$ acceptance and
$3.86\times$ speedup.

\subsection{Effect of Additional Training}
\label{sec:sft_ablation}

To verify that the improvement of ReTrace does not simply result from
additional training, we construct an SFT baseline by continuing to train
DFlash with its original block-diffusion objective~\cite{chen2026dflash},
using the same training data and optimization budget as ReTrace. The only
difference is that the SFT baseline does not condition on rejected-suffix
representations.

\begin{figure}[t]
    \centering
    \includegraphics[width=\columnwidth]
    {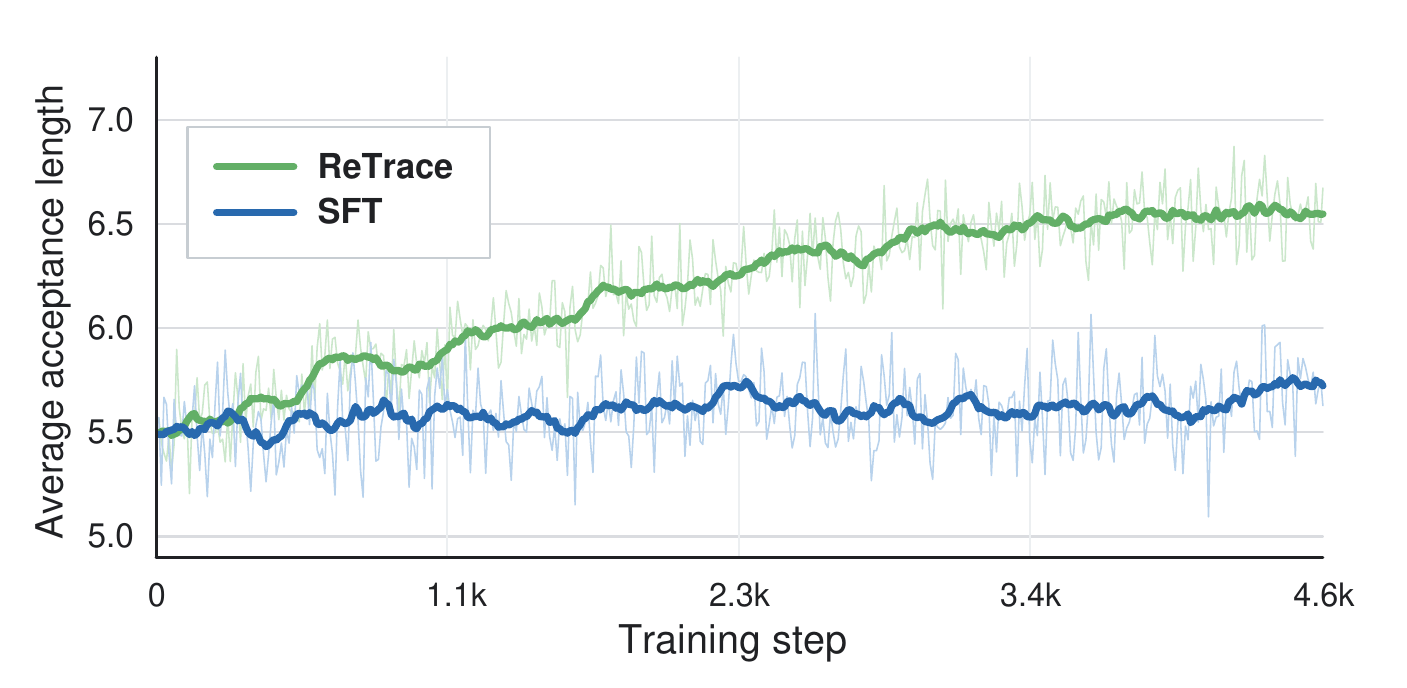}
    \caption{
    Average acceptance length during training. The SFT baseline continues
    training with the original DFlash objective using the same data and
    training budget as ReTrace, but without rejected-suffix conditioning.
    }
    \label{fig:sft_ablation}
\end{figure}

As shown in Figure~\ref{fig:sft_ablation}, continued DFlash training improves
only marginally and quickly saturates, whereas ReTrace steadily increases the
average acceptance length throughout training, indicating that the gain comes
from conditioning on the rejected suffix rather than from additional training.

\subsection{Runtime Efficiency Analysis}
\label{sec:runtime_overhead}

We measure the per-cycle latency of DFlash and ReTrace on GSM8K with Qwen3-4B on a single NVIDIA A800 GPU, using context length 1,024 and batch size 1. We report median CUDA-event latency, and the total is the sum of all non-overlapping components in one decoding cycle.
\begin{table}[t]
\centering
\setlength{\tabcolsep}{8pt}
\renewcommand{\arraystretch}{1.06}
\resizebox{\columnwidth}{!}{%
\begin{tabular}{lcc}
\toprule
Metric & DFlash & ReTrace \\
\midrule
Verify latency (ms)          & $23.56$ & $23.58$ \\
Draft latency (ms)           & $8.34$  & $8.38$  \\
LM-head latency (ms)         & $0.59$  & $0.59$  \\
Conditioning overhead (ms)   & --      & $0.13$  \\
Total latency (ms)           & $32.49$ & $32.68$ \\
\midrule
Speedup                      & $3.37\times$ & $\mathbf{3.73\times}$ \\
Avg.\ accepted length $\tau$ & $6.36$ & $\mathbf{7.36}$ \\
\bottomrule
\end{tabular}%
}
\caption{Runtime efficiency on GSM8K with Qwen3-4B. The conditioning overhead combines target-aware correction and gated fusion.}
\label{tab:runtime_efficiency}
\end{table}
As shown in Table~\ref{tab:runtime_efficiency}, ReTrace raises the speedup from $3.37\times$ to $3.73\times$ and the average accepted length from $6.36$ to $7.36$, at a combined correction and fusion cost of $0.13$ ms per cycle, only $0.40\%$ of the DFlash cycle latency. It requires neither an additional target-model forward pass nor an autoregressive drafting step.

\subsection{Compatibility with Domino}
\label{sec:domino_compatibility}

Domino improves draft quality by modeling causal dependencies within the current draft block, whereas ReTrace conditions each block on the rejected suffix of the previous round, so the two methods optimize orthogonal information pathways.

\begin{table}[t]
\centering
\resizebox{\columnwidth}{!}{
\begin{tabular}{lcccc}
\toprule
Benchmark
& \multicolumn{2}{c}{Throughput (tok/s)}
& \multicolumn{2}{c}{Acceptance Length} \\
\cmidrule(lr){2-3}
\cmidrule(lr){4-5}
& Domino
& +ReTrace
& Domino
& +ReTrace \\
\midrule
MATH-500
& 343.28
& \textbf{356.61} {\scriptsize(+13.33)}
& 9.22
& \textbf{9.55} {\scriptsize(+0.33)} \\

AIME25
& 278.24
& \textbf{288.94} {\scriptsize(+10.70)}
& 6.85
& \textbf{7.17} {\scriptsize(+0.32)} \\

LiveCodeBench
& 247.79
& \textbf{266.71} {\scriptsize(+18.92)}
& 6.76
& \textbf{7.22} {\scriptsize(+0.46)} \\

HumanEval
& 279.38
& \textbf{293.30} {\scriptsize(+13.92)}
& 6.88
& \textbf{7.23} {\scriptsize(+0.35)} \\
\midrule
Macro Average
& 287.17
& \textbf{301.39} {\scriptsize(+14.22)}
& 7.43
& \textbf{7.79} {\scriptsize(+0.36)} \\
\bottomrule
\end{tabular}
}
\caption{Compatibility with Domino on selected math and code benchmarks. All results use greedy decoding and a maximum generation length of 1,024 tokens. Absolute improvements over Domino are shown in parentheses, and bold values indicate the better result.}
\label{tab:domino_compatibility}
\end{table}

As shown in Table~\ref{tab:domino_compatibility}, adding ReTrace to Domino raises macro-average throughput from $287.17$ to $301.39$ tok/s and mean acceptance length from $7.43$ to $7.79$ tokens on the selected math and code benchmarks, confirming that cross-round conditioning is compatible with Domino's intra-block causal refinement.

\FloatBarrier

\section{Conclusion}

We revisit the rejected suffix in speculative decoding and show that, although
its tokens cannot be committed, its hidden representations retain predictive
information that benefits subsequent drafting. Based on this observation, we
introduce ReTrace, which conditions each draft block on the rejected suffix of
the previous round: the retained states are aligned, corrected with target-side
signals, and admitted through gated residual fusion. Across model scales,
decoding temperatures, and diverse benchmarks, ReTrace consistently improves
acceptance length and end-to-end speedup over its DFlash backbone at negligible
latency cost, and remains compatible with intra-block improvements such as
Domino. Rejected draft computation therefore need not be wasted: it can condition
the drafting that follows.

\bibliographystyle{plainnat}
\bibliography{main}

\clearpage
\beginappendix
\setcounter{secnumdepth}{2}

\section{Algorithm}
\label{appendix:algorithm}

Algorithm~\ref{alg:retrace} summarizes one ReTrace decoding round. Lines
4--6 are the only additions to the underlying speculative-decoding loop: they
correct the signal carried from the previous round and admit it into the draft
input. Everything else, including target verification and sampling, is
unchanged. The conditioning signal is reset whenever a block is fully accepted
or the rejection occurs at the last block position, in which case the drafter
falls back to its original behavior.

\begin{algorithm}[h]
\caption{One ReTrace decoding round}
\label{alg:retrace}
\textbf{Input}: context $x$, target model $\mathcal{M}_T$, drafter $\mathcal{M}_D$, block size $B$ \\
\textbf{Parameter}: $W_c,W_v,W_g$, correction coefficient $\beta$ \\
\textbf{Output}: extended context $x$
\begin{algorithmic}[1]
\STATE $m_{1,j}\leftarrow\mathbf{0}$, $u_{1,j}\leftarrow\mathbf{0}$ for $1\leq j\leq B$
\FOR{$t=1,2,\ldots$ until EOS}
\STATE $e_{t,0:B}\leftarrow$ drafter inputs (anchor and $B$ proposal positions)
\STATE $c_{t,j}\leftarrow W_c[m_{t,j};u_{t,j}]$ \hfill $\triangleright$ Eq.~\ref{eq:memory_correction_update}
\STATE $r_{t,j}\leftarrow m_{t,j}+\beta\,c_{t,j}$ \hfill $\triangleright$ Eq.~\ref{eq:corrected_memory}
\STATE $\widetilde{e}_{t,j}\leftarrow e_{t,j}+W_v r_{t,j}\odot\sigma\!\left(W_g[e_{t,j};r_{t,j}]\right)$ \hfill $\triangleright$ Eq.~\ref{eq:gated_conditioning}
\STATE $\hat{y}_{t,1:B},\,h^{D}_{t,1:B}\leftarrow\mathcal{M}_D(\widetilde{e}_{t,0:B})$
\STATE $A_t,\,h^{T}_{t,1:B}\leftarrow\textsc{Verify}(\mathcal{M}_T,x,\hat{y}_{t,1:B})$
\STATE append the $A_t$ accepted tokens and the correction token to $x$
\IF{$A_t<B-1$}
\STATE $r_t\leftarrow A_t+1$ \hfill $\triangleright$ first rejected position
\STATE $m_{t+1,j}\leftarrow h^{D}_{t,r_t+j}$ and $u_{t+1,j}\leftarrow h^{T}_{t,r_t+j-1}$ for $1\leq j\leq B-r_t$
\STATE $m_{t+1,j}\leftarrow\mathbf{0}$ and $u_{t+1,j}\leftarrow\mathbf{0}$ for $B-r_t<j\leq B$
\ELSE
\STATE $m_{t+1,j}\leftarrow\mathbf{0}$, $u_{t+1,j}\leftarrow\mathbf{0}$ \hfill $\triangleright$ nothing to carry
\ENDIF
\ENDFOR
\STATE \textbf{return} $x$
\end{algorithmic}
\end{algorithm}

\section{Training Data}
\label{appendix:training_data}

ReTrace is trained on a prompt-only pool of 40K examples drawn from
Open-PerfectBlend, an open reproduction of the instruction mixture
of~\cite{xu2024perfectblend}, which spans dialogue, mathematical reasoning, code
generation, and instruction following. The pool is built with random seed $42$.

\section{Model and ReTrace Configuration}
\label{appendix:model_config}

We use Qwen3-4B and Qwen3-8B as target models and initialize their drafters from
the corresponding public DFlash-b16 checkpoints, each of which has five draft
layers and a block size of $16$. The target model remains frozen, and all target
representations used for correction are taken from the ordinary verification
pass, so no additional target forward pass is required.
Table~\ref{tab:appendix_model_config} lists the configuration.

\begin{table}[h]
\centering
\small
\setlength{\tabcolsep}{4pt}
\begin{tabular}{lcc}
\toprule
Setting & Qwen3-4B & Qwen3-8B \\
\midrule
Hidden size (target, drafter) & 2,560 & 4,096 \\
Drafter layers              & 5 & 5 \\
Draft block size            & 16 & 16 \\
\midrule
Reused representation       & \multicolumn{2}{c}{Final draft-layer state} \\
Correction representation   & \multicolumn{2}{c}{Final target-layer state} \\
Conditioning lifetime       & \multicolumn{2}{c}{One drafting round} \\
Retained states             & \multicolumn{2}{c}{FP16, no gradient} \\
Maximum $\beta$             & \multicolumn{2}{c}{1.0} \\
Correction warmup           & \multicolumn{2}{c}{Linear, 50 iterations} \\
\bottomrule
\end{tabular}
\caption{Model and ReTrace configuration.}
\label{tab:appendix_model_config}
\end{table}

\section{Optimization}
\label{appendix:optimization}

Training uses the same block-diffusion objective as the DFlash
backbone~\cite{chen2026dflash}. No additional target-model loss, counterfactual
gate loss, or auxiliary conditioning loss is introduced: the correction and gated
fusion parameters are optimized through the current-round prediction loss alone.
The draft backbone is optimized jointly with them, while the retained states are
detached between rounds. Training runs on 32
NVIDIA A800 40GB GPUs across four eight-GPU nodes. Table~\ref{tab:appendix_optimization}
lists the hyperparameters.

\begin{table}[h]
\centering
\small
\setlength{\tabcolsep}{4pt}
\resizebox{\columnwidth}{!}{%
\begin{tabular}{ll}
\toprule
Training hyperparameter & Value \\
\midrule
Initialization              & DFlash-b16 checkpoint \\
Precision                   & BF16 \\
Optimizer                   & AdamW, weight decay 0.01 \\
Learning rate               & $5\times10^{-5}$, cosine decay \\
LR warmup ratio             & 0.05 \\
Prompt / response length    & 512 / 1,024 tokens \\
Prompt batch size           & 32 \\
Training epochs             & 8 \\
Distributed strategy        & FSDP \\
Seed                        & 42 \\
\bottomrule
\end{tabular}%
}
\caption{Optimization hyperparameters.}
\label{tab:appendix_optimization}
\end{table}

The correction coefficient $\beta$ of Equation~\ref{eq:corrected_memory} is
phased in with a linear warmup. Writing $\beta_s$ for its value at training
iteration $s$,
\begin{equation}
\beta_s
=
\beta_{\max}
\min\!\left(
\frac{s}{T_{\mathrm{warm}}},
1
\right),
\qquad
T_{\mathrm{warm}}=50 ,
\label{eq:correction_warmup}
\end{equation}
so correction is inactive at the start of training and is introduced gradually.
At inference time the fully warmed value $\beta=\beta_{\max}=1$ is used. Since
$W_v$ is zero-initialized, ReTrace exactly reproduces the original draft input
before learning begins.

\section{Evaluation Protocol}
\label{appendix:evaluation}

Within each comparison, all methods use the same prompts, target checkpoints,
tokenizer, stopping criteria, and sampling parameters, and thinking mode is
disabled throughout. Unless an experiment states otherwise, evaluation issues one
request at a time with a draft block size of $16$ for DFlash and ReTrace, the
candidate tree size indicated in parentheses for EAGLE-3, and at most $8{,}192$
new tokens, terminating earlier on EOS. We evaluate $128$ prompts each on GSM8K,
MATH-500, LiveCodeBench, and Alpaca, and the full test sets of HumanEval
($164$), AIME25 ($30$), and MT-Bench ($80$).

At temperature $0$, greedy decoding uses \texttt{top\_p}$=1.0$ and
\texttt{top\_k}$=1$; at temperature $1$, sampling uses \texttt{top\_p}$=0.95$ and
\texttt{top\_k}$=20$. The random seed is $42$. Before measuring a benchmark we
issue one warmup batch and clear the runtime cache. End-to-end speedup is
computed against vanilla autoregressive decoding of the same target model under
the same generation configuration. Average acceptance length $\tau$ is computed
over target verification cycles. Latency includes proposal generation, ReTrace
conditioning, target verification, sampling, and cache maintenance, and excludes
model loading and the benchmark warmup.

\subsection{Runtime Microbenchmark}

The runtime decomposition reported in the main paper is measured on one NVIDIA
A800 GPU with Qwen3-4B, batch size $1$, context length $1{,}024$, greedy
decoding, and draft block size $16$. Component times are measured with CUDA
events and reported as medians. The total is the sum of the non-overlapping
verification, draft, LM-head, and conditioning components of one
speculative-decoding cycle.

\section{Ablation Settings}
\label{appendix:ablation_settings}

The component ablation uses Qwen3-4B, block size $16$, greedy decoding, and the
same checkpoint initialization, training data, optimization budget, and
evaluation prompts for every row. Raw conditioning, overwrite replaces the
active mask inputs with the aligned rejected states; raw conditioning,
gated retains the mask pathway and applies only gated residual fusion; and
corrected conditioning, gated adds target-aware correction before the same
gated fusion.

The SFT baseline of Figure~\ref{fig:sft_ablation} continues DFlash training under
the original block-diffusion objective with the same training data and
optimization budget as ReTrace, differing only in that it does not condition on
rejected-suffix representations. For the Domino compatibility study,
all methods use greedy decoding, block size $16$, and a maximum generation length
of $1{,}024$ tokens; Domino operates within a draft block, while ReTrace carries
information only between consecutive blocks.

\section{Concurrent-Serving Experiment}
\label{appendix:concurrency}

The main results measure one request at a time. We additionally evaluate
concurrent serving, where speculative decoding must compete with the batching
that already saturates the target model.

The experiment uses Qwen3-4B and the first $128$ GSM8K test prompts, with
no-think greedy decoding, BF16, FlashInfer, and at most $1{,}024$ new tokens.
DFlash and ReTrace use a $16$-token draft block; EAGLE-3 uses seven draft steps,
top-$k=4$, and tree sizes $16$ and $60$. We sweep client concurrency over
$c\in\{1,2,4,8,16,32\}$, running each method in an isolated SGLang server on one
NVIDIA A800-SXM4-40GB GPU and measuring all five configurations simultaneously.
The server captures CUDA Graphs for all batch sizes up to $32$, including the
post-verification correction and fusion path. At each concurrency level we flush
the cache and issue one warmup wave of $c$ requests before timing. ReTrace here
uses the deployment-efficient low-rank realization of its conditioning path:
rank-$64$ rejected-suffix latents, input-side latent residual fusion, rank-$32$
target-aware correction, and a one-round conditioning lifetime.

\begin{center}
\begin{minipage}{\columnwidth}
\centering
\setlength{\tabcolsep}{2pt}
\renewcommand{\arraystretch}{1.06}
\footnotesize
\resizebox{\columnwidth}{!}{%
\begin{tabular}{ccccc}
\toprule
\multirow{2}{*}{Conc.} & EAGLE-3 & EAGLE-3 & DFlash & ReTrace \\
 & (16) & (60) & (16) & (16) \\
\midrule
1  & $1.67\times$ / 3.31 & $1.67\times$ / 3.70 & $3.37\times$ / 6.36 & $\mathbf{3.73\times}$ / \textbf{7.36} \\
2  & $1.55\times$ / 3.31 & $1.54\times$ / 3.70 & $3.20\times$ / 6.42 & $\mathbf{3.49\times}$ / \textbf{7.38} \\
4  & $1.47\times$ / 3.31 & $1.24\times$ / 3.69 & $2.89\times$ / 6.43 & $\mathbf{3.15\times}$ / \textbf{7.43} \\
8  & $1.31\times$ / 3.31 & $0.90\times$ / 3.69 & $2.38\times$ / 6.38 & $\mathbf{2.56\times}$ / \textbf{7.36} \\
16 & $1.07\times$ / 3.31 & $0.61\times$ / 3.71 & $1.83\times$ / 6.37 & $\mathbf{2.01\times}$ / \textbf{7.34} \\
32 & $0.80\times$ / 3.31 & $0.42\times$ / 3.69 & $1.50\times$ / 6.45 & $\mathbf{1.62\times}$ / \textbf{7.42} \\
\bottomrule
\end{tabular}%
}
\captionof{table}{Concurrent serving on GSM8K with Qwen3-4B. Each entry reports the
end-to-end speedup over target-only decoding at the same client concurrency and
the average acceptance length $\tau$. Bold indicates the best result at each
concurrency level.}
\label{tab:concurrent_serving}
\end{minipage}
\end{center}

As shown in Table~\ref{tab:concurrent_serving}, ReTrace is the fastest method at
every concurrency level and gains about one token of acceptance length over
DFlash. The advantage survives batching because the conditioning signal is
reused inside the drafter and adds no verification work. At concurrency $32$,
ReTrace still reaches $1.62\times$ while both EAGLE-3 configurations fall below
target-only decoding. The concurrency-one row reproduces the corresponding
single-request measurements of Table~\ref{tab:main-results}.


\end{document}